# Distribution Assertive Regression

Kumarjit Pathak [a*] , Jitin Kapila [b*], Aasheesh Barvey [c], Nikit Gawande [d]

[a] Data Scientist professional , Harman , Whitefield, Bangalore ,mail:Kumarjit.pathak@outlook.com

[b] Data Scientist professional , Zeta Global , Indiranagar, Bangalore ,mail:Jitin.kapila@outlook.com

[c] Data Scientist professional , Harman , Whitefield, Bangalore ,mail:ashbarvey@gmail.com

[b] Data Scientist professional , Northwestern University , ,mail: nikitgawande2018@u.northwestern.edu

**Abstract:** In regression modelling approach, the main step is to fit the regression line as close as possible to the target variable. In this process most algorithms try to fit all of the data in a single line and hence fitting all parts of target variable in one go. It was observed that the error between predicted and target variable usually have a varying behavior across the various quantiles of the dependent variable and hence single point diagnostic like MAPE has it's limitation to signify the level of fitness across the distribution of Y(dependent variable). To address this problem, a novel approach is proposed in the paper to deal with regression fitting over various quantiles of target variable. Using this approach we have significantly improved the eccentric behavior of the distance (error) between predicted and actual value of regression.

Our proposed solution is based on understanding the segmented behavior of the data with respect to the internal segments within the data and approach for retrospectively fitting the data based on each quantile behavior. We believe exploring and using this approach would help in achieving better and more explainable results in most settings of real world data modelling problems.

## INTRODUCTION

Most of value estimation/prediction problems in the industry are solved by regression or by classification (this includes clustering and association too) techniques. Regression analysis follows the concept of drawing functions of a line / plane / boundary, which is as close as possible to the target / dependent variable. Using this function, we try to interpolate or extrapolate the values so that we can reasonably tell the possibility of target / dependent value given independent datum.

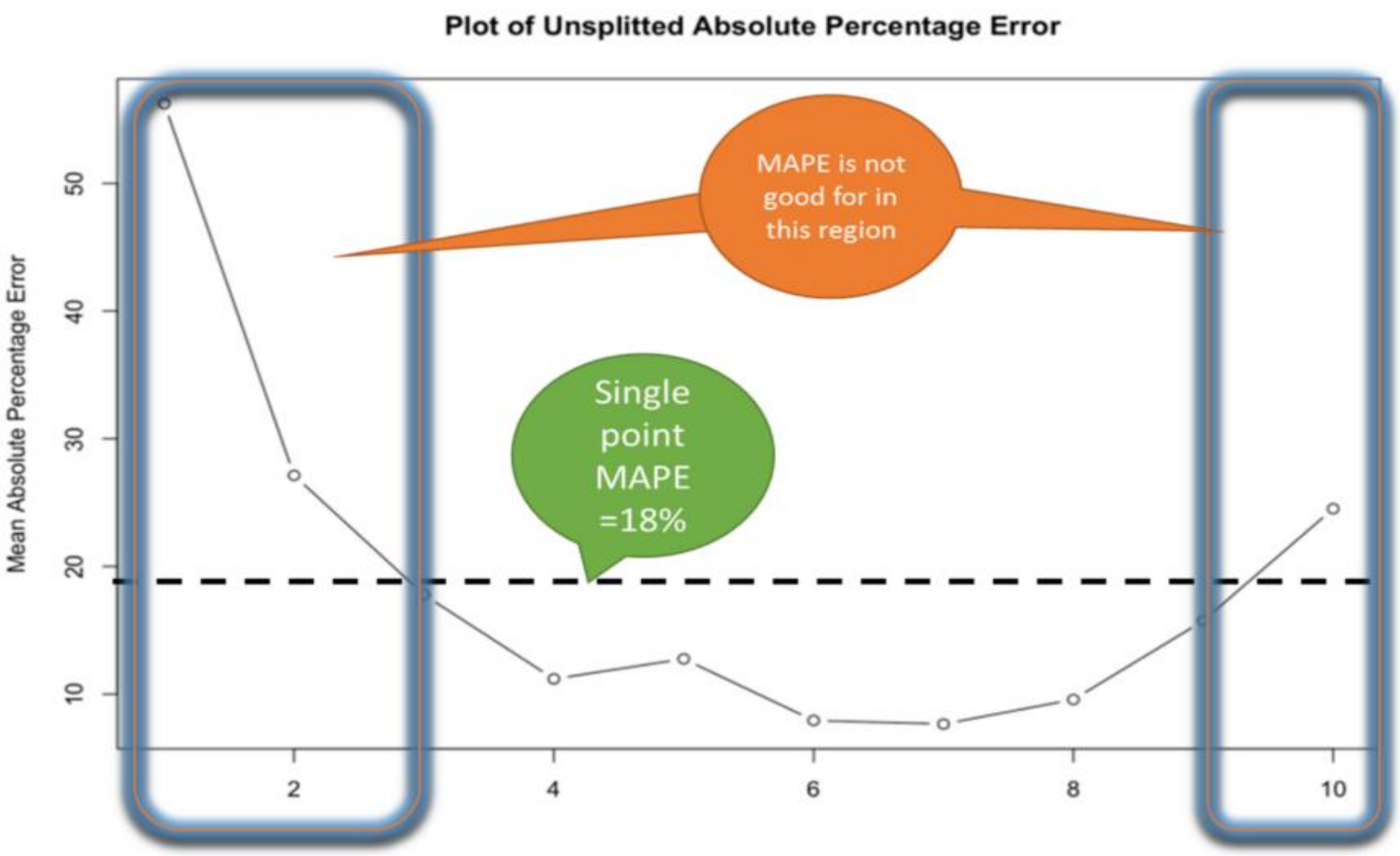

Fig. 1. Mean Absolute Percentage Error plot at each quantile of the dependent variable using "BostonHousing" data [11].

As we can observe that drawing this line function (from this point onwards referred as "line function" as synonymous to Regression

“line” / “plane” / “boundary”) from the data is key challenge to solve the regression problem. Data Scientists experiments on regression using OLS, MLE, Ridge, LASSO, Robust etc., and evaluated them using RMSE (Root Mean/Median Square Error), MAD (Mean/Median Absolute Deviation), MAE (Mean / Median Absolute Error) and MAPE (Mean/Median Absolute Percentage Error) et al [1],etc.

But all of these gives a single point estimate(single value) that what is the overall error looks like. However it is not easy to infer that our trained model has fitted well across the distribution of dependent variable? If certain region is not fitted, well we might need to look at the same to see what might have gone wrong or do we need to build a separate model to represent behavior of the subpart of the data.

With our approach, we are able to partially segment the data and are able to signify a new datum’s reference w.r.t the partial segment. This not only enhances the explanation power of line function but also makes it more robust to outlier behavior within the training data or with new datum flowing in.

## I. Literature Survey

Regression technique is existing since 200 years, there are lot of development has happened on the same from time to time to address different issues and advancement required based on the data complexity increase over time along with demand of analytics based decision making.

Previously to solve regression problem many research has been followed ordinary least squares to maximum likelihood estimators to robust estimators and many research has been done with using Bayesian methods, quantile, quadratic, least-angle and principal component, neural networks and mixed model approach. An exhaustive list of regression provided on Wikipedia [1], and survey of estimators, are available as well.

Recently in adaptive regression, MARS (Multivariate Adaptive Regression Splines) et al [2],[12] and ANN (Artificial Neural Networks) [12] have approached the non-linearity in a very effective manner and were able to achieve significant success in understanding the impact of independent variable on dependent variable. Most of these approaches look at line functions as a function of varying beta coefficients of independent variable that impacts the dependent variable. The main problem what we observe is that, they up to certain extent assume that the target variable follows a single distribution or has homogeneous behavior across the data.

Trees approach like CART (Classification and Regression Trees) et al [3], Bagging Trees et al [4] and Random Forests et al [5], approaches line function as occurrences of segment of line outcome. In a way, they tend to do a very high multi-class classification on target variable where classes are the grouped segments of the target variable. These have significant robustness towards outlier and understands non-linearity up to great extent. The main shortcoming of this approach is that it loses the interpretability and it cannot give continuous output as the other regression methods can offer.

Similarly the similarity search approaches like KNN et al [7] and Associative Curvature et al [8], provide very nice results but as tree behaves like multi-class classification problem and tends to fail on interpretability.

In our approach we try to fill in the gaps between these methodologies by taking a simple approach of *“split and perform”*. The framework proposed by us tries to bridge the best of the above approaches by utilizing the addictiveness, data understanding, robustness and interpretability of these approaches.

All current evaluation and performance metrics of fitting a regression line is based on a point estimator, like MAPE which gives just one value. Similarly we have RMSE, MAD, MAP, R-squared, Adj. R-squared et al [1],etc. However in practice it was observed that this is not sufficient metric to explain the overall behavior of fitment as there can be still possibilities that in some part or segment of target variable the fitment is not good enough compared to the other values.

In this paper we intend to propose a solution this above mentioned problem.

## II. Method and Procedure

Our framework of *"split and perform"* works on the concept of *bathtub curve*. In actual practice, while estimating the line function, we use MAPE (Mean Absolute Percentage Error) as one of the methods of estimating how close or far is the values from line function is with actual target value.

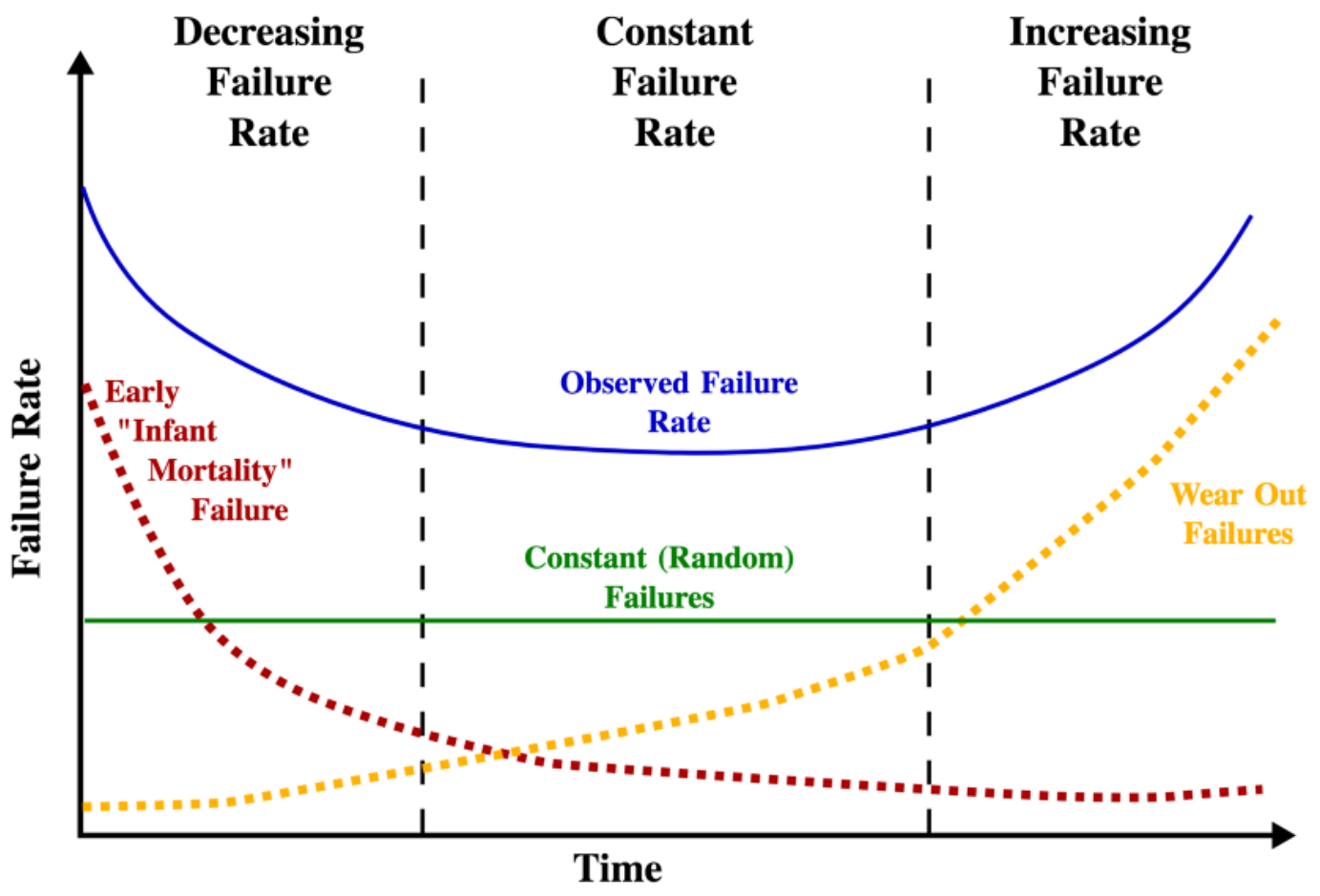

Fig. 2. Bathtub curve [12]

**MAPE:**

The beauty of this simple measure is that it tells how far are we from actual value in terms of percentage. But the downside is it averages out the whole range percentage error and does not imply the internal change of this value. We here exploits this beauty of MAPE but we enhance the shortcomings by fitting this over *deciles* of the target variable.

Following this practice we observe a Bathtub curve [10] with most of line functions we used to fit on same data. As we know from the field of Reliability Engineering, a bathtub curve can be broken into three parts: Initial High Failure, Stable Zone, Wear Out Zone. Understanding this behavior and applying it to MAPE of Deciles of Target Values we infer the following:

1. The models usually performs best towards the central part of the regression, but performs poorly towards both ends of target variable.
2. Since we tend to optimize the line functions over majority of dependent variables, the outlier behavior within the data causes noise at both ends of the data.
3. As from reliability engineering standpoint, the Initial Failure Rate for our line functions is the inability to classify data that belongs to heavy lower range of data. At same time Wear Out Region is the models inability to classify higher extremes outlier behavior within the data.
4. The Stable Zone for line functions is the middle range of target variables.

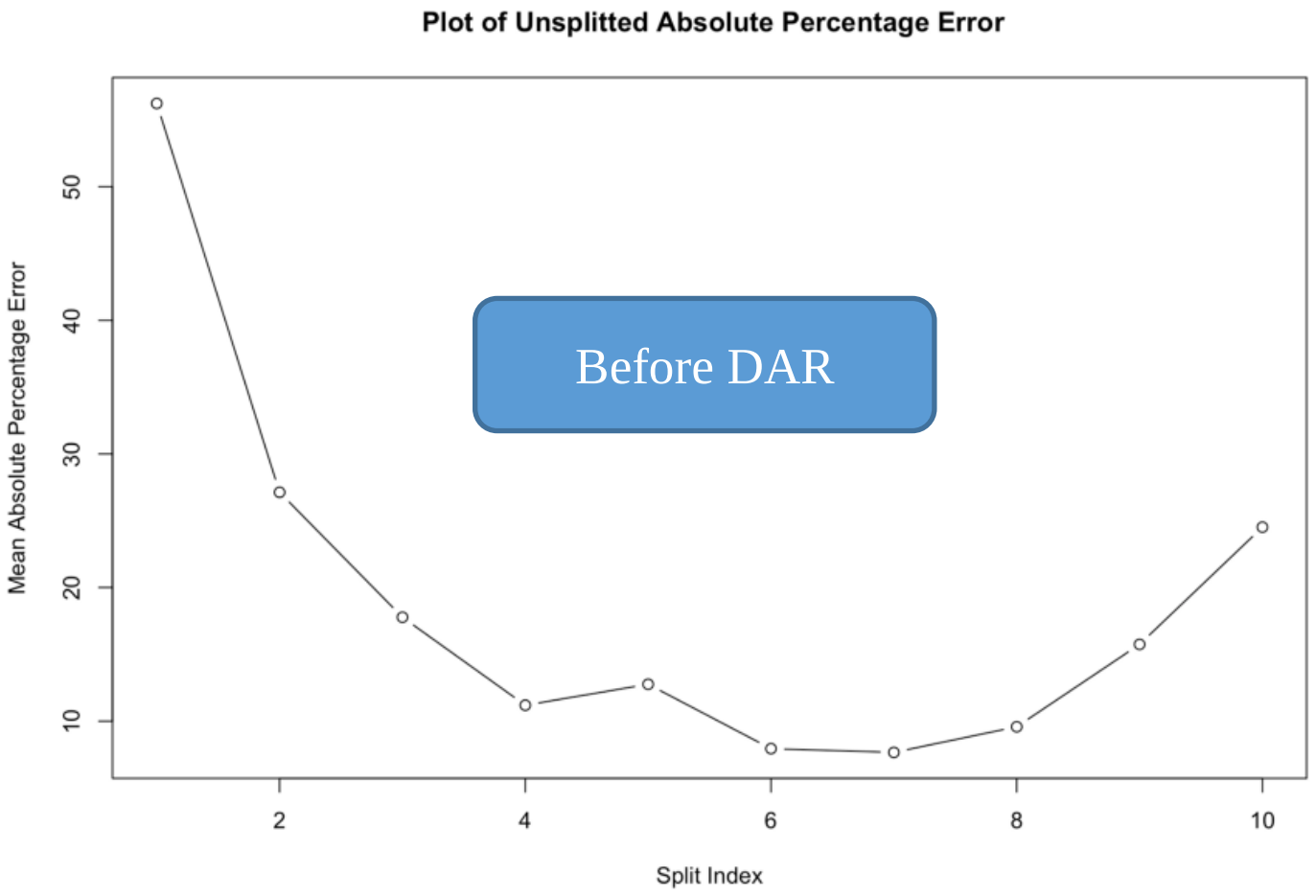

Fig. 3. MAPE at each quantile of the dependent variable before applying our approach

However, we don't have a proper review of this behavior across huge set of algorithms and datasets to standardize the limit value of how many value but to our observation on real data the Initial Failure Rate exist for first 2-3 deciles whereas Wear Out Range is for 8-10 deciles of target variable..

The study data for the paper has been taken for UCI library et al.[11] and plot line shows behavior of various datasets. What is particular about these results is that, they are more or less same on various data, affirming the first observation stated above. Now to tweak this a bit we then added additional outlier values to data and ran on the models the behavior confirms to our observation stated in point 2 and 3.

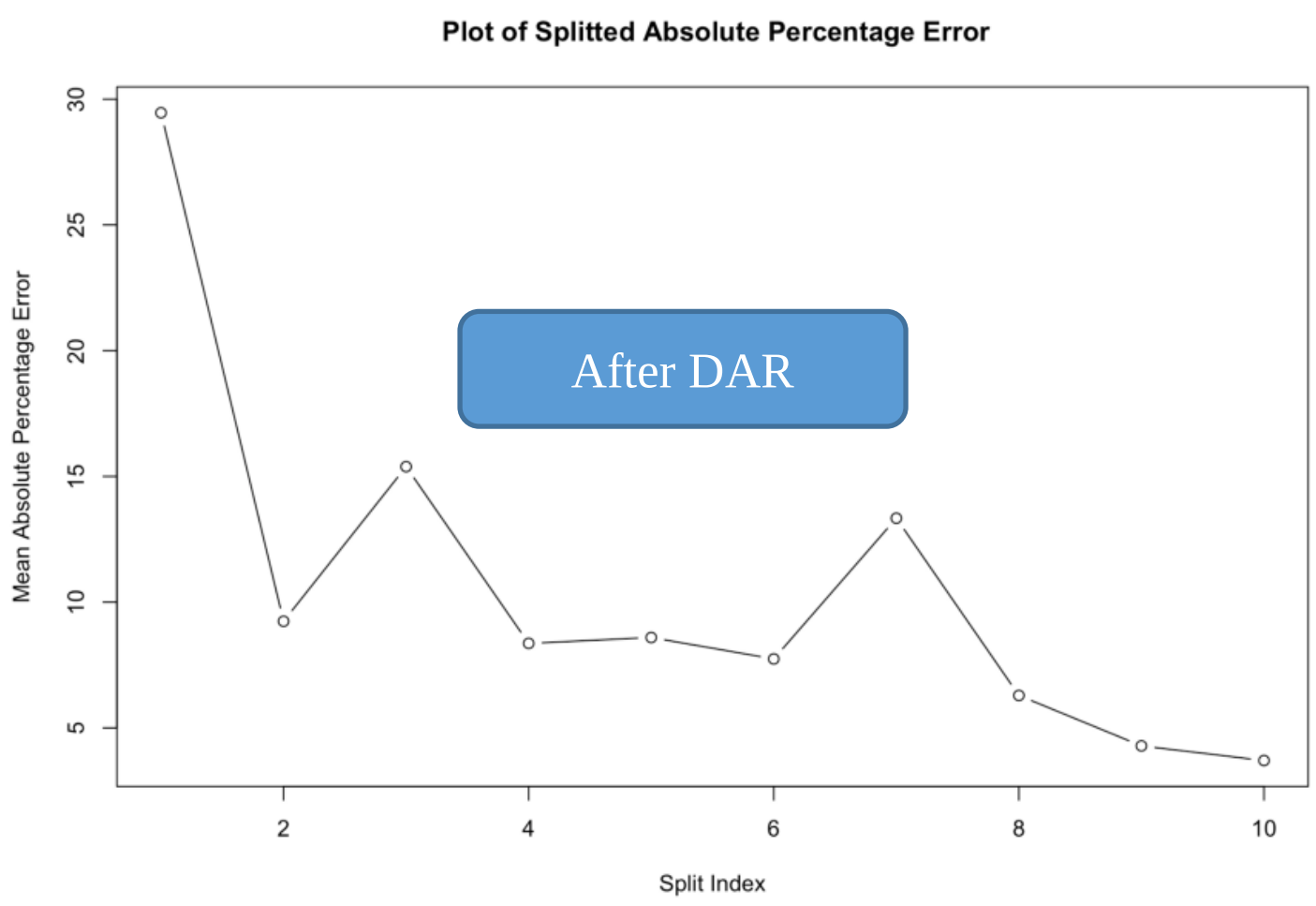

Fig. 4. MAPE plot at each quantile after applying our approach

To test our observation we added random noise to the data which belongs to the middle section and found the model to perform stably within the limits. The observation of above tests as showcased in Fig. 4

From observations we understand that if we are able to segment the data we will be in a better position to score the new data with much reasonable accuracy.

Now to do this we propose the framework where we can fit in the new line functions. The as from the *bathtub* curve observed above we split the data into three sub-groups / segments based on the behavior in the curve and make three different model on the same segments' data. Now while scoring we find whether the new data belongs to which of these segment and score them with the model of that segment. The Algorithm is defined in DAFR flow below.

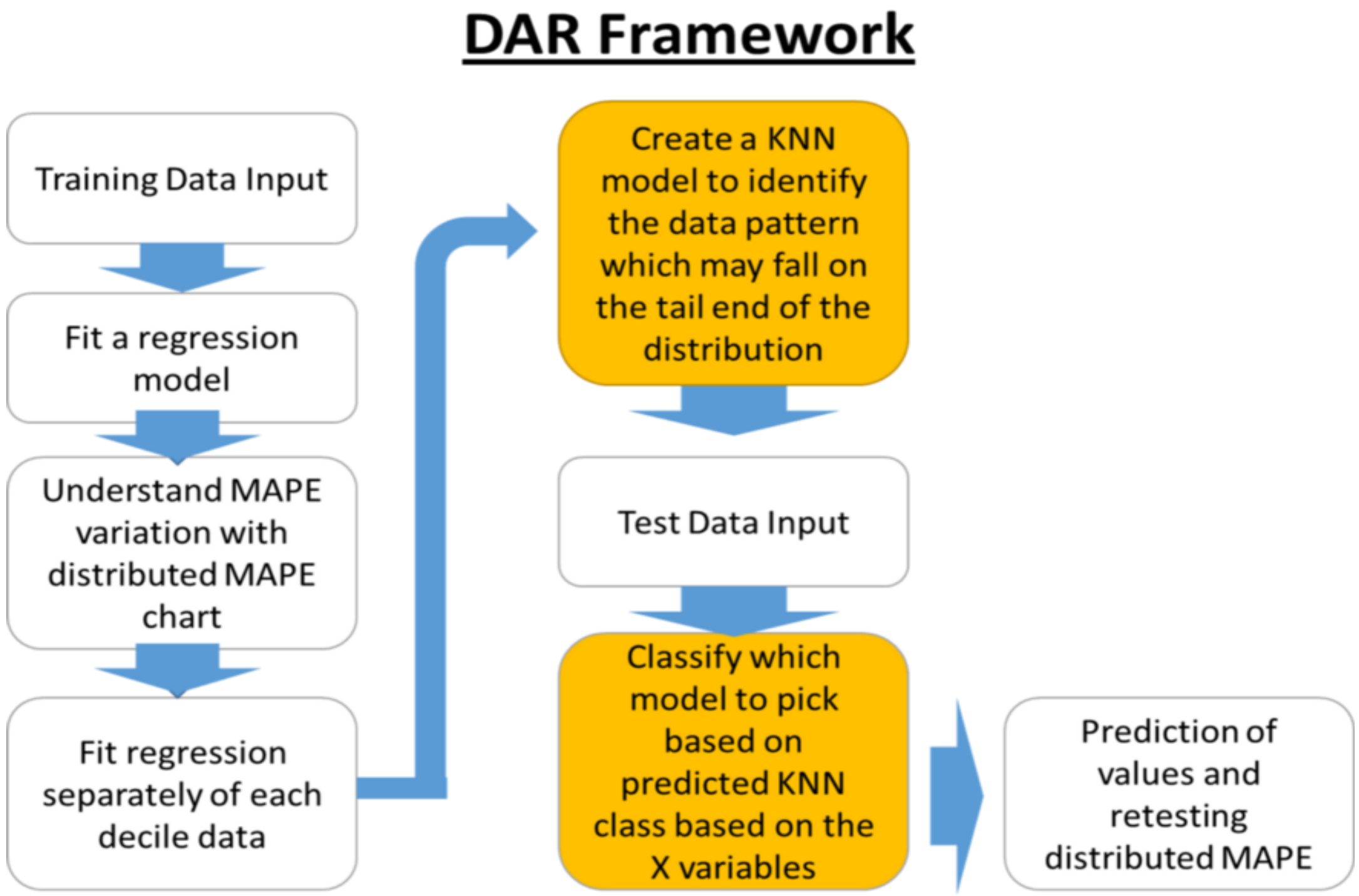

Fig. 4. MAPE plot at each quantile after applying our approach

Also we need to understand that this method can be extensively used with all above algorithms stated in literature review, but in our practice we found simple Maximum Likelihood Estimate Regression as line function and KNN as Similarity function works well. We chose this selection as we don't want to lose on interpretability of the data.

**Algorithm of Distribution Assertive Regression:**

---

**Algorithm 1**: Deca - Adaptive Fit Regression Training
**Require:** Fit Function (Fn) and Similarity Function (Sn)
**Step 1:** Train all data with Fn with X for Y
**Step 2:** Score all data with Fn to give Yhat
**Step 3:** Calculate MAPE of Deciles with (Y,Yhat)
**Step 4:** Split data based on MAPE in X_front, X_mid, and X_back
**Step 5:** Train Fn_front, Fn_mid, Fn_back with X_front, X_mid, X_back
**Step 6:** Score on Fn_front, Fn_mid, Fn_back to get Yn_front, Yn_mid, Yn_back
**Step 7:** Calculate MAPE with Yn_front, Yn_mid, Yn_back with Y
**Step 8:** Based On Sn keep train for similarity
**Step 9:** Return Fn_front, Fn_mid, Fn_back and Sn
**End**

---

Post training we will score new data from following algorithm:

---

**Algorithm 1**: Scoring Deca - Adaptive Fit Regression
**Require:** Fit Function (Fn), Similarity Function (Sn), New Data Xnew
**Step 1:** Score with Sn to find Segment of Xnew
**Step 2:** Based on Segment of Xnew score with Fn_front, Fn_mid, Fn_back as Ynew
**Step 3:** Return Ynew
**End**

---

## III. Conclusion

From our study and analysis, we believe that this extensible and flexible framework can be used to train regression models in a much better way. Also, as we can see the behavior on the deciles and segment the data accordingly we can observe the intrinsic behaviour of the data as front, mid and back segments. The presence of these segments makes the model more robust to outlier and hence more capable to handle any unforeseen data

## IV. Future Scope

We believe that this framework is very flexible and extensible, but we foresee following possible improvements:

1) Similarity search can be improved using problisitic scoring models.
2) Similarity weighted scoring from an line functions
3) Gradient change based automatic segment selection

4) Variable Impact based Segment selection
5) Singe Decile based modelling
6) Tree based similarity search

Though this not an exhaustive list but these might be possible paradigms to explore this framework.